# Class Specific Feature Selection for Interval Valued Data through Interval K-Means Clustering

D S Guru and N Vinay Kumar

Department of Studies in Computer Science, University of Mysore, Manasagangotri, Mysore-570006, Karnataka, INDIA
dsg@compsci.uni-mysore.ac.in, vinaykumar.natraj@gmail.com

**Abstract.** In this paper, a novel feature selection approach for supervised interval valued features is proposed. The proposed approach takes care of selecting the class specific features through interval K-Means clustering. The kernel of K-Means clustering algorithm is modified to adapt interval valued data. During training, a set of samples corresponding to a class is fed into the interval K-Means clustering algorithm, which clusters features into K distinct clusters. Hence, there are K number of features corresponding to each class. Subsequently, corresponding to each class, the cluster representatives are chosen. This procedure is repeated for all the samples of remaining classes. During testing the feature indices correspond to each class are used for validating the given dataset through classification using suitable symbolic classifiers. For experimentation, four standard supervised interval datasets are used. The results show the superiority of the proposed model when compared with the other existing state-of-the-art feature selection methods.

## 1 Introduction

In the current era of digital technology- pattern recognition plays a vital role in the development of cognition based systems. These systems quite naturally handle a huge amount of data. While handling such vast amount of data, the task of data processing has become curse to process. To overcome curse in data processing, the concept of feature selection is being adopted by researchers. Nowadays, feature selection has become a very trending topic in the field of machine learning and pattern recognition. Feature selection helps us to select the most relevant features from a given set of features. The different feature selection techniques can be listed as: filter, wrapper, and embedded methods [1].

Generally, the existing conventional methods fail to perform feature selection on unconventional data like interval, multi-valued, modal, and categorical data. These data are also called in general symbolic data. The notion of symbolic data was emerged in the early 2000, which mainly concentrates in handling very realistic type of data for effective classification, clustering, and even regression for that matter [2].

As it is a powerful tool in solving realistic problems, we thought of developing a feature selection model for any one of the modalities. In this regard, we have chosen with an interval valued data, due its strong nature in preserving the continuous streaming data in discrete form [2]. In this regard, we built a feature selection model for interval valued data in this work.

In literature, works done on feature selection of interval valued data is very few compared to conventional feature selection methods. Ichino [3] had given only the theoretical interpretation of feature selection on interval valued based on the pretended simplicity algorithm which works in a Cartesian space. Bapu et. al., [4] proposed a two stage feature selection algorithm which can handle both interval as well multi-valued data based on the Mutual Similarity Value proximity measure. But the method is restricted to unsupervised dataset. Lyamine et. al., [5] speak about the feature selection of interval valued data based on the concept of similarity margin computed between an interval sample and a class prototype. The similarity margin is computed using certain symbolic similarity measure. The authors have constructed basis for the similarity margin and then they worked out at the multi-variate weighting scheme. The weights guarantee with the number of subset features to be selected. The experimentation is done on three standard benchmarking interval dataset and validated using LAMBA classifier. Recently, Chih-Ching et. al., [6] have come up with a feature selection model, where it adopts the framework of [5] for selecting the features. But, the authors have used different kernel viz., Gaussian kernel for measuring the similarity between an interval sample and a class prototype. The authors also have given the experimentation and comparative analysis on only single interval valued dataset.

Apart from the above mentioned works, no work can be seen on feature selection of interval valued data based on clustering in general, class specific features selection in particular. The class specific feature selection helps in improving the prediction accuracy as it selects features which are most relevant to the specific classes instead of selecting features relevant to all classes. With this motivational background, here in this paper, a feature selection model is proposed and validated on supervised interval datasets.

The proposed feature selection model initially transforms the given supervised feature matrix and later divides the transformed matrix into several (equal to number of classes) interval feature sub-matrices. The transformed feature sub-matrices are then fed into interval K-means clustering algorithm. Thus, results with K clusters for each sub-matrix. Next, for every cluster, a cluster representative is computed that results with K number of representatives corresponding to each class. During testing, a single interval feature vector is considered for classification based on the class specific features selected from the knowledgebase and they are validated using suitable symbolic classifiers.

The major contributions of this paper are as follows:

1. Proposal of a novel interval feature selection model based on class specific feature selection.
2. Design of an interval K-Means clustering algorithm by incorporating interval similarity kernel into conventional K-Means clustering.

3. Conduction of an experimentation, to show that the proposed model outperforms the state-of-the art models.

The organization of the paper is as follows: In section 2, the proposed model is well explained. The details of experimental setup, datasets and results are given in section 3. Section 4 presents a comparative analysis. Finally, section 5 concludes the paper.

## 2  Proposed Model

The proposed interval feature selection model comprises of various steps in both training and testing respectively. Pre-processing, Interval K-Means clustering, selection of cluster representatives (Feature Selection) are done at a former stage and selection of feature indices (pre-processing), and classification tasks are performed at a latter stage. The architecture of the proposed model is given in Figure 1.

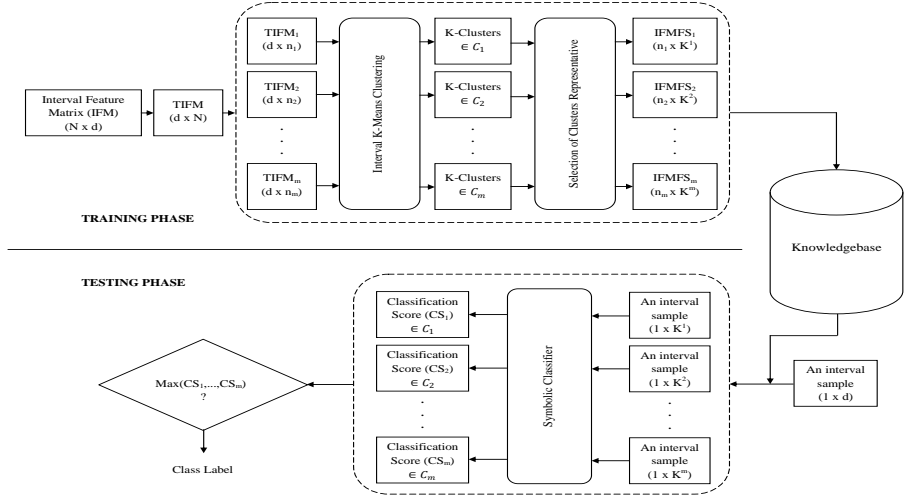

**Fig. 1.** Architecture of the Proposed Model

### 2.1  Pre-processing during training phase

Let us consider a supervised interval feature matrix IFM, with N number of rows and d+1 number of columns. Each row corresponds to a sample and each column corresponds to a feature of type interval. Such a matrix is given by: IFM: $(\xi_i, y_i)$, $\xi_i$ represents a sample and $y_i$ represents a class label ($i = 1, 2, ..., N$).

Each sample $\xi_i$ is described by d interval features and is given by:

$$\xi_i = \left(\xi_i^1, \xi_i^2, ..., \xi_i^d\right) = ([f_1^-, f_1^+], [f_2^-, f_2^+], ..., [f_d^-, f_d^+])$$

where, $f_k^-$ and $f_k^+$ are the lower and upper limits of an interval respectively.

The basic idea of the proposed model is to cluster the similar features. To accomplish this, a feature matrix should be transformed in such a way that the positions of samples become features and features become samples. That is, the rows of a feature matrix should correspond to features and the columns should correspond to samples.

Now, we have a transposed feature matrix of dimension $d \times N$. But, our main objective is to select class specific features. In this regard, we separated samples based on their class correspondence and obtain with a sub-matrix $TIFM_j$ of dimension $d \times n_j$ ($n_j$ is the number of samples per class) corresponding to a particular class $C_j$ ($j = 1, 2, \ldots, m; m = no. of\ classes$).

Further, the $TIFM_j$ is fed into interval K-Means clustering algorithm to obtain K clusters from each matrix, where $d$ features are spread across K different clusters. The details of clustering procedure are given in next section.

### 2.2 Interval K-Means Clustering

In this section, details about the construction of an interval K-Means clustering algorithm are given.

As we know, conventional K-Means clustering is an instance of partitional clustering techniques. Initially, it fixes up with the number of clusters (K) and the centroid points. Then the algorithm uses one of the different kernels such as squared Euclidean/ city block/ cosine/ correlation/ Hamming distance to compute the proximity among the samples [7]. Later those samples which have greater affinity go to same cluster and samples with a little affinity go to different clusters. Then the new centroids will be computed for each K clusters. The same procedure is repeated until certain convergence criteria are satisfied. The convergence criteria may be the maximum iterations or $\epsilon$ difference. Usually, the above said procedures are followed in conventional K-Means clustering algorithm. But, in our work, as we are handling with interval valued data, a slight modification has been brought out at kernel level. The kernel used to compute the affinity among the samples is symbolic similarity measure [8] instead of the above said kernels. The symbolic similarity kernel (SSK) used in our work is given by:

$$SSK(A,B) = \frac{(ISV_{AB}^- + ISV_{AB}^+) + (ISV_{BA}^- + ISV_{BA}^+)}{4} \quad (1)$$

Where A = $([a_1^- a_1^+], [a_2^- a_2^+], \ldots, [a_k^- a_k^+], \ldots)$ and B = $([b_1^- b_1^+], [b_2^- b_2^+], \ldots, [b_k^- b_k^+], \ldots)$ be any two interval objects with $(a_k^- \leq a_k^+)$ and $(b_k^- \leq b_k^+)$. $ISV_{AB}^-, ISV_{AB}^+ (ISV_{BA}^-, ISV_{BA}^+)$ be the lower and upper limits of interval similarity value computed from object A to object B (B to A) and are given by:

$$ISV_{AB}^- = min(Sim(A_k, B_k)); k = 1,2,\ldots,d$$

$$ISV_{AB}^+ = max(Sim(A_k, B_k)); k = 1,2,\ldots,d$$

$$ISV(A,B) = [ISV^-, ISV^+]$$

$$Sim(A_k, B_k) = \begin{cases} 1 & \text{if A contains B} \\ \dfrac{\text{Length of overlapping portion of A and B}}{\text{Length of B}} & \text{If there exists overlapping} \\ 0 & \text{If there is no overlapping} \end{cases}$$

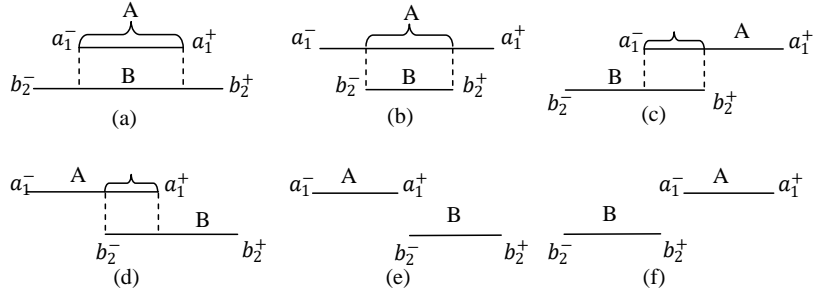

**Fig. 2.** Different instances of similarity computation between two intervals [4], (a)-(d): Overlapping cases, (e)-(f): Non-overlapping cases

The symbolic similarity kernel used in this work is very realistic in nature, as it preserves the topological relationship between the two intervals. The illustration of different cases of topological relationships exist between two intervals are shown in Figure 2.

The rest of the interval K-Means clustering follows the same procedure as conventional K-Means clustering as explained above.

### 2.3 Selection of Cluster Representative (Feature Selection)

The K-clusters obtained from the class are considered here for selecting respective clusters' representative. The procedure for selecting the cluster representatives is discussed below:

Consider a cluster $Cl_q \in C_j$ (q = 1,2,…,K; j = 1,2,…,m), containing $z$ number of features. A feature is said to be a cluster representative $(ClR_q)$, then it must exhibit a maximum similarity to all the remaining features. In this regard, the similarity computation among the features in a cluster $Cl_q$ results with a similarity matrix $SM^q$, which is given by:

$$SM_{ab}^q = SSK(F_a, F_b) \in \mathbb{R}, \forall a = 1,2,…,z; b = 1,2,…,z$$

Where, $SSK(F_a, F_b)$ is given by equation (1).

Further, the computation of average similarity value is given by:

$$ASV_a = \frac{1}{z}\sum_{b=1}^{z} SM_{ab}^q \qquad (2)$$

Now, we have obtained with $z$ average similarity values corresponding to cluster $Cl_q$. Thus $ClR_q$ is given by:

$$ClR_q = \arg\max_{F_a}\{(ASV_1, ASV_2, \ldots, ASV_z)'\} \quad (3)$$

$$i.e., ClR_q = \arg\max_{F_a}\left\{\left(\frac{1}{z}\sum_{b=1}^{z} SM_{1b}^q, \frac{1}{z}\sum_{b=1}^{z} SM_{2b}^q, \ldots, \frac{1}{z}\sum_{b=1}^{z} SM_{zb}^q\right)'\right\}$$

From equation 3, it is very clear that the feature ($F_a$) which gives maximum $ASV_a$ value is considered as a cluster representative $(ClR_q)$ corresponding to cluster $Cl_q$. Thus the above procedure is repeated for all the clusters corresponding to remaining classes. Now we have mxK such clusters representative $(ClR_1, ClR_2, \ldots, ClR_{mK})'$. The feature indices of these cluster representatives are further used to select features from original interval feature matrices and archived the same in the knowledgebase for classification. The illustration of selection of cluster representatives (features selection) is shown in Figure 3.

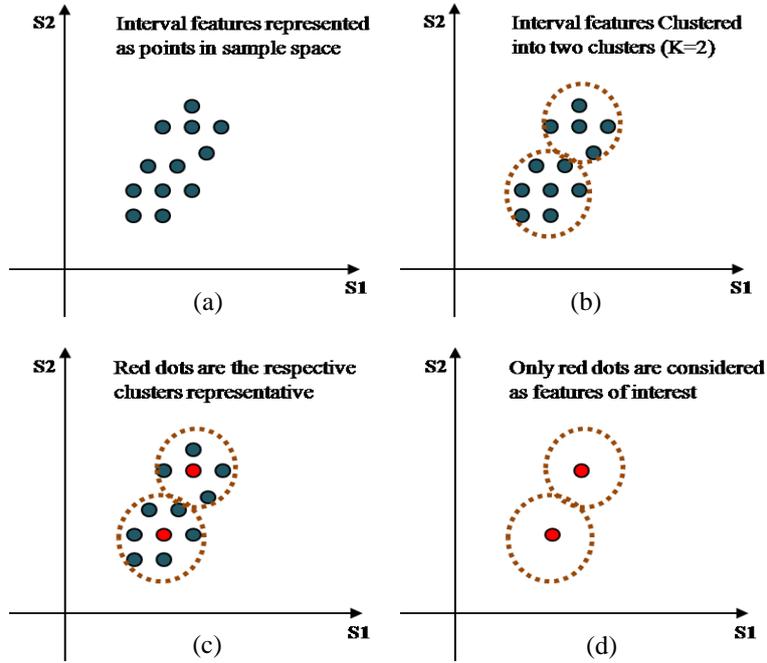

**Fig. 3.** Illustration of interval features selection using clustering: (a) Interval features (visualized as points) are spread in two dimensional sample space, (b) Clusters obtained after applying interval K-Means clustering, (c) Red dots are the clusters representative selected based on the procedure explained in section 2.3, (d) Two red dots are further retained as the features of interest.

### 2.4 Pre-processing during testing phase

Basically, our classification strategy is influenced from biometric verification, where, initially it identifies a class of the sample and then authenticates the claimed sample during verification. Similarly, in our work, initially a query sample is claimed as not only the member of a single class instead it is claimed to be as the member of all m different classes. While claiming the membership of all m classes, the (K) features of a query sample are selected based on the feature indices preserved during training. Thus for classification, there are totally m different query instances (of a sample) entering into the symbolic classifiers.

### 2.5 Symbolic Classifiers

In this paper, as we are handling with interval type data, it is difficult to compatible with conventional classifiers such as K-NN (K-Nearest Neighbour), SVM (Support Vector Machines), Random Forest etc [9]. In this regard, we have recommended symbolic classifiers which handle interval type data successfully for classification. Here, we use two symbolic classifiers proposed in [10], and [11]. Henceforth, these two are called as C-1 and C-2 classifiers respectively.

The classifier proposed in [10], operates directly on interval data and the similarity measure proposed by them is of interval type in nature, hence the computed similarity matrix is again a symbolic. Further, authors aggregate the obtained matrix using the concept of mutual similarity value (MSV) and obtain with the conventional symmetric similarity matrix. Later on, the authors follow the nearest neighbour approach for classification.

In [11], authors propose a symbolic classifier which mainly concentrates on nearest neighbour approach in classifying an unknown sample to a known class. In our work, a slight modification has been done on the similarity measure proposed by [11]. The slight modification of the above said similarity measure is done at feature level. As the similarity measure in [11], is capable of measuring the similarity of multi-interval valued data, but in our case, only single interval valued data is enough to measure the similarity between two objects. Hence, we have restricted the measure to only single interval valued data.

Finally, we end up with $m$ different classification scores obtained for $m$ different classes. The classification of an unknown sample is labelled based on these scores. A class which possesses a maximum classification score is given as the class membership for an unknown sample.

## 3 Experimentation and Results

### 3.1 Datasets

We have used totally four different supervised interval datasets for experimentation. The four different benchmarking datasets used are: Iris [2], Car [12], Water [13], and Fish [12] datasets. The Iris interval dataset consists of 30 samples with 4 interval features. The 30 samples are spread across the three different classes, with 10 samples per class. The Car dataset consists of 33 samples with 8 interval features. The samples are spread across the 4 different classes with 10, 8, 8, and 7 samples per class respectively. The Fish dataset consists of 12 samples with 13 interval features. The samples are spread across 4 different classes, each with 4, 2, 4, and 2 samples respectively. Finally, the Water dataset consists of 316 samples with 48 interval features. The samples are spread across 2 different classes, each with 223 and 93 samples respectively. Among four datasets, the fish dataset seems to provide very less instances for classification. Thus for this dataset, the classification results vary a lot compare to other datasets.

### 3.2 Experimental Setup

In this sub-section, details of experimentation conducted on the four benchmarking supervised interval datasets are given. The experimentation is conducted in two phases viz., training, and testing. During training phase, we consider a supervised interval feature matrix and obtained the class specific features as explained in section 2. These features are then preserved in a knowledgebase for classification. During testing phase, an unknown interval sample is considered and selected its class specific features and performed classification as explained in sections 2.4 and 2.5.

During training and testing, the samples of the dataset are varied from 30 percent to 70 percent (in steps 10 percent) and 70 percent to 30 percent (in steps 10 percent) respectively. For, interval K-Means clustering the maximum number of iterations is fixed to be 100. The value of K in interval K-Means clustering is varied from 2 to one less than number of features for all datasets except water dataset (In case water dataset, the K is varied till 21, as the clustering algorithm does not converge above that value). The parameter $\beta$ in the symbolic classifier (C-2) is fixed to be 1.

### 3.3 Results

The validation of the proposed feature selection model is performed using classification accuracy, defined as the ratio of correctly classified samples to the number of samples. The experimental results are tabulated for the best feature subset (results with feature selection (WFS)) obtained from the proposed model and also we have compared the classification results of the same datasets without using any feature selection models (results without feature selection (WoFS)). The tabulated results are shown from Table 1 to Table 4.

**Table 1.** Comparison of classification accuracies obtained from the classifiers C-1 and C-2 with different training-testing percentage for Iris dataset

| Train-Test | C-1 [10] | | | | C-2 [11] | | | |
|---|---|---|---|---|---|---|---|---|
| | WFS | | WoFS | | WFS | | WoFS | |
| | # of Features | Accuracy | # of Features | Accuracy | # of Features | Accuracy | # of Features | Accuracy |
| 30-70 | 3 | 100 | 4 | 100 | 3 | 100 | 4 | 100 |
| 40-60 | 2 | 100 | 4 | 100 | 2 | 100 | 4 | 100 |
| 50-50 | 2 | 100 | 4 | 100 | 3 | 100 | 4 | 100 |
| 60-40 | 2 | 100 | 4 | 91.67 | 2 | 100 | 4 | 100 |
| 70-30 | 2 | 100 | 4 | 88.89 | 3 | 100 | 4 | 100 |

**Table 2.** Comparison of classification accuracies obtained from the classifiers C-1 and C-2 with different training-testing percentage for Car dataset

| Train-Test | C-1 [10] | | | | C-2 [11] | | | |
|---|---|---|---|---|---|---|---|---|
| | WFS | | WoFS | | WFS | | WoFS | |
| | # of Features | Accuracy | # of Features | Accuracy | # of Features | Accuracy | # of Features | Accuracy |
| 30-70 | 2 | 76.19 | 8 | 42.86 | 2 | 61.90 | 8 | 71.43 |
| 40-60 | 4 | 72.22 | 8 | 38.89 | 4 | 61.11 | 8 | 77.78 |
| 50-50 | 3 | 81.25 | 8 | 50 | 4 | 56.25 | 8 | 81.25 |
| 60-40 | 2 | 66.67 | 8 | 50 | 4 | 83.33 | 8 | 83.33 |
| 70-30 | 2 | 66.67 | 8 | 55.56 | 2 | 55.56 | 8 | 88.89 |

**Table 3.** Comparison of classification accuracies obtained from the classifiers C-1 and C-2 with different training-testing percentage for Fish dataset

| Train-Test | C-1 [10] | | | | C-2 [11] | | | |
|---|---|---|---|---|---|---|---|---|
| | WFS | | WoFS | | WFS | | WoFS | |
| | # of Features | Accuracy | # of Features | Accuracy | # of Features | Accuracy | # of Features | Accuracy |
| 30-70 | 3 | 66.67 | 13 | 66.67 | 6 | 50 | 13 | 66.67 |
| 40-60 | 5 | 100 | 13 | 66.67 | 7 | 66.67 | 13 | 66.67 |
| 50-50 | 10 | 66.67 | 13 | 66.67 | 10 | 66.67 | 13 | 66.67 |
| 60-40 | 3 | 100 | 13 | 100 | 2 | 50 | 13 | 100 |
| 70-30 | 8 | 100 | 13 | 100 | 2 | 50 | 13 | 100 |

**Table 4.** Comparison of classification accuracies obtained from the classifiers C-1 and C-2 with different training-testing percentage for Water dataset

| Train-Test | C-1 [10] | | | | C-2 [11] | | | |
|---|---|---|---|---|---|---|---|---|
| | WFS | | WoFS | | WFS | | WoFS | |
| | # of Features | Accuracy | # of Features | Accuracy | # of Features | Accuracy | # of Features | Accuracy |
| 30-70 | 11 | 68.78 | 48 | 66.52 | 2 | 71.50 | 48 | 63.80 |
| 40-60 | 14 | 68.09 | 48 | 65.96 | 14 | 71.81 | 48 | 60.11 |
| 50-50 | 5 | 70.70 | 48 | 65.61 | 6 | 73.89 | 48 | 56.05 |
| 60-40 | 4 | 69.84 | 48 | 54.76 | 20 | 77.78 | 48 | 62.70 |
| 70-30 | 2 | 76.34 | 48 | 63.44 | 10 | 79.57 | 48 | 59.14 |

From the above tables (Table 1 to 4), it is very clear that the classification performance has been increased due to the incorporation of feature selection method compared to that of not using any feature selection method. It is also so clear that the best results are quoted for lesser number of features.

To test effectiveness of the proposed model, we have used the datasets containing different feature dimensions (from 4 to 48). From the results, one can notice that the model performs well even for such kind of datasets in-spite of selecting very few features. This shows the robustness of the proposed model under varied feature sizes of the dataset.

## 4 Comparative Analysis

To corroborate the effectiveness of the proposed model, the comparative analyses are given. Initially, the proposed model is compared against the state-of-the-art methods in terms of classification accuracy and the same is tabulated in Table 5. Further, we have compared our model with the other models which do not use any feature selection during classification. In literature, we found such classification models reported only on Car dataset. Hence, we have given comparison with only Car dataset and is given in Table 6.

**Table 5.** Comparison of proposed feature selection method v/s other existing methods

|  | Lyamine et. al., [5] | | Chih-Ching et. al., [6] | | **Proposed model** | | | |
|---|---|---|---|---|---|---|---|---|
| Classifier | LAMDA | | LAMDA | | C-1 [10] | | C-2 [11] | |
| Dataset | Feature Subset | Accuracy | Feature Subset | Accuracy | Feature Subset | Accuracy | Feature Subset | Accuracy |
| Car | 5 | 78 | -- | -- | 3 | 81.25 | 4 | **83.33** |
| Fish | 4 | 74 | -- | -- | 3 | **100** | 7 | 66.67 |
| Water | 14 | 77 | 11 | 78.66 | 2 | 76.34 | 10 | **79.57** |

The Table 6 justifies that the classification model with feature selection outperforms well compared to that of other models which do not use any feature selection methods during classification. This is because; the proposed feature selection model selects only the features of interest through clustering instead of considering all features.

From Table 5 and Table 6, it is very clear that the proposed model not only outperforms the existing models in terms of accuracy but also in terms of selecting number of features. This shows that the proposed model is much better than the similarity margin based models [5] and [6]. The selected feature subset and their class specific features indices corresponding to C-1 and C-2 classifiers are given in Table 7.

**Table 6.** Comparison of classification with the proposed feature selection method v/s the existing classification methods without using any feature selection methods for Car interval dataset

| Methods | | Accuracy |
|---|---|---|
| **Proposed Method** | C-1 (WFS) | **81.25** |
| | C-2 (WFS) | **83.33** |
| Barros et. al., [14] | Binary Model | 48.49 |
| | Multinomial | 54.55 |
| Renata et. al., [15] | IDPC-CSP | 63.64 |
| | IDPC-VSP | 63.6 |
| | IDPC-PP | 72.8 |
| Silva and Bruto, [16] | Distributional approach | 73 |
| | Mid points and Ranges | 55 |

Table 7. Selected feature subsets and their corresponding feature indices

| Dataset (# of Class) | C-1 | C-2 |
|---|---|---|
| Car (4) | {(1,3,4), (2,4,1), (3,1,2), (1,8,4)} | {(4,3,6,1), (1,2,3,4), (2,1,5,4), (1,8,6,5)} |
| Fish (4) | {(12,4,7), (8,1,9), (4,12,1), (9,3,6)} | {(1,9,3,12,2,4,7), (1,3,5,9,11,6,7), (9,11,1,5,8,6,12), (3,12,2,9,8,13,7)} |
| Water (2) | {(5,18), (23,34)} | {(38,42,18,8,34,35,40,24,4,14), (13,32,27,31,22,34,36,18,3,16)} |

## 5  Conclusion

In this paper, a novel idea for selection of supervised interval data through clustering is introduced. The proposed model incorporates the concept of symbolic similarity measure to build the interval K-Means clustering. The cluster representatives are computed based on the symbolic similarity measure. Later on, the indices of cluster representatives are preserved in the knowledgebase. During testing, for a sample, the class specific feature indices are selected from knowledgebase and classified using symbolic classifier. The proposed model has been well exploited for different interval supervised datasets and also it outperformed against other existing models in terms of classification accuracy and also in terms of dimension.